\documentclass{bmvc2k}
\usepackage{times}
\usepackage{epsfig}
\usepackage{graphicx}
\usepackage{amsmath}
\usepackage{amssymb}
\usepackage{caption}
\usepackage{multirow}
\usepackage{arydshln}
\usepackage{color}
\usepackage[table,xcdraw]{xcolor}

\graphicspath{{./images/}}

\title{Semi-Supervised Raw-to-Raw Mapping}

\addauthor{Mahmoud Afifi}{https://www.eecs.yorku.ca/~mafifi/}{0}
\addauthor{Abdullah Abuolaim}{https://www.eecs.yorku.ca/~abuolaim/}{0}

\addinstitution{
Lassonde School of Engineering\\
York University\\
Toronto, Ontario\\
Canada
}

\runninghead{Afifi and Abuolaim}{Semi-Supervised Raw-to-Raw Mapping}

\def\mappingloss{$\mathcal{L}_m$}
\def\anchorloss{$\mathcal{L}_a$}
\def\recloss{$\mathcal{L}_r$}

\def\mappinglossEQ{\mathcal{L}_m}
\def\anchorlossEQ{\mathcal{L}_a}
\def\reclossEQ{\mathcal{L}_r}

\begin{document}

\maketitle

\begin{figure}[h]
\includegraphics[width=\textwidth]{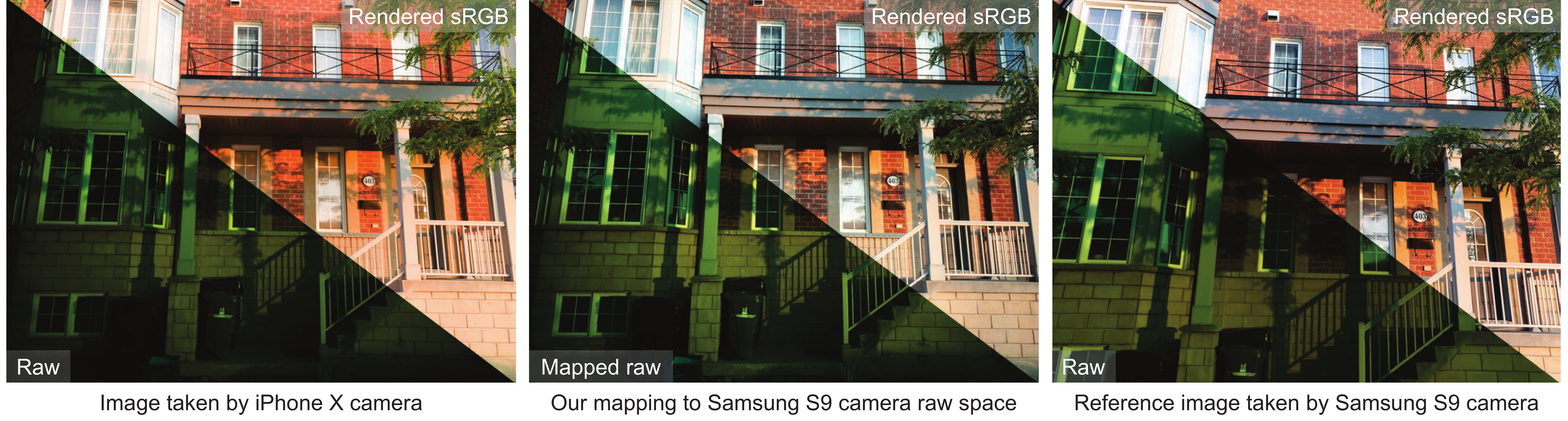}
\vspace{-1mm}
\caption{We introduce a semi-supervised raw-to-raw mapping method. This figure shows a raw image captured by iPhone X smartphone camera and our mapping result to Samsung Galaxy S9 smartphone camera raw space, along with a reference image capturing the same scene by Samsung Galaxy S9. For each image, we show both raw and the camera-ISP rendered image. Shown images are from our dataset. Note that the original Bayer raw image is packed to $\texttt{RGGB}$ channels. To aid visualization, the Green channels are averaged (i.e., three-channel $\texttt{RGB}$ image), and a Gamma operation with $1/1.6$ encoding gamma is applied. This is applicable for raw visualization in the rest of this paper.}
\label{fig:teaser}
\end{figure}

\begin{abstract}
The raw-RGB colors of a camera sensor vary due to the spectral sensitivity differences across different sensor makes and models. 
This paper focuses on the task of mapping between different sensor raw-RGB color spaces. Prior work addressed this problem using a \textit{pairwise} calibration to achieve accurate color mapping. Although being accurate, this approach is less practical as it requires: (1) capturing pair of images by both camera devices with a color calibration object placed in each new scene; (2) accurate image alignment or manual annotation of the color calibration object. This paper aims to tackle color mapping in the raw space through a more practical setup. Specifically, we present a semi-supervised raw-to-raw mapping method trained on a small set of paired images alongside an unpaired set of images captured by each camera device. Through extensive experiments, we show that our method achieves better results compared to other domain adaptation alternatives in addition to the single-calibration solution. We have generated a new dataset of raw images from two different smartphone cameras as part of this effort. Our dataset includes unpaired and paired sets for our semi-supervised training and evaluation. 
\end{abstract}

\section{Introduction and Related Work} \label{sec:intro}

A camera image signal processor (ISP) applies a set of operations to render captured images from the camera's internal raw space into a standard display color space (e.g., standard RGB, or sRGB for short). While colors in the sRGB space can be different across cameras, it is also possible to observe color differences in the camera raw space when capturing the same scene by different camera devices. The reason behinds these color differences can be understood from the raw image formation. Mathematically, the construction of a raw image, $I = \{I_r, I_g, I_b\}$, can be described as follows \cite{basri2003lambertian}: 
\begin{equation}
\label{eq0}
I_c(x) =\int_{\gamma} \rho(x,\lambda)R(x,\lambda)S_{c}(\lambda) d\lambda,
\end{equation}
\noindent where $c = \{\text{R}, \text{G}, \text{B}\}$, $x$ refer to pixel location in $I$, and $\gamma$ and $\rho(\cdot)$ are the visible light spectrum and the spectral power distribution of the scene illumination, respectively. The captured object spectral reflectance properties are denoted by $R(\cdot)$, and $S(\cdot)$ refers to the camera sensor sensitivity at wavelength $\lambda$. Note that we omit image noise in Eq. \ref{eq0} for simplicity, as our focus in this paper is mainly related to raw colors.

From Eq. \ref{eq0}, it is clear that if camera $A$ and camera $B$ capture the same scene to produce images $I_A$ and $I_B$ (assuming they are perfectly aligned), the final colors in $I_A$ and $I_B$ would be similar \textit{iff} the camera sensor sensitivities, $S$, of camera $A$ and $B$ are the same. Typically, this case rarely happens, especially when different vendors manufactured the camera sensors. As a result, $I_A$ and $I_B$ are likely to have different colors even when capturing the same scene under the same lighting conditions \cite{jiang2013space}. It is also clear that differences in colors produced by cameras $A$ and $B$ may have different levels of variations across scenes/lighting conditions due to the integral of  $\rho(\cdot)R(\cdot)S_{c}(\cdot)$ on the visible light spectrum, $\gamma$, in Eq. \ref{eq0}.

%  It is also clear that differences in colors produced by cameras $A$ and $B$, when capturing the same scene under the same lighting, may have different levels of variations across scenes due to the integral of  $\rho(\cdot)R(\cdot)S_{c}(\cdot)$ on the visible light spectrum, $\gamma$, in Eq. \ref{eq0}.

Raw-to-raw mapping aims to reduce color differences in $I_A$ and $I_B$, and it is useful for camera ISP manufacturing as elaborated next. First, raw-to-raw mapping is useful for data generation in any camera ISP learning-based module that utilizes raw images. For instance, prior work \cite{lou2015color, fourure2016mixed, afifi2021ciexyz, afifi2020cross} showed that color mapping for data augmentation purposes could improve the accuracy of color constancy in raw images. %However, prior work did not address raw-to-raw mapping explicitly.

Second, camera ISPs include different carefully calibrated modules inherently tied to the camera sensor space used in designing such modules. When a camera manufacturer introduces a new sensor with a different spectral sensitivity, these tuned camera ISP modules should be adapted to the spectral sensitivity of this new sensor \cite{ignatov2020replacing, schwartz2018deepisp, liang2021cameranet}. Needless to say, this adaptation process often requires collecting new labeled data with some corresponding ground-truth annotations. This process is tedious, and, as a result, deploying a new sensor in a camera device is still challenging and requires a lot of human effort.

To avoid generating new labeled data when employing a new sensor, recent work in \cite{liba2019handheld} proposes to map color histogram of new sensor raw images to the original sensor space used to train the illumination estimation camera ISP module. Then, the estimated illumination color is projected back to the new sensor space. 
Building on top of the idea in \cite{liba2019handheld}, one could either: (1) map labeled training ISP images from the old sensor space to the new sensor space, and thus the retraining is practicable for all learning-based ISP modules; or (2) design a ``universal camera ISP'' by mapping all images to this specific sensor space. Fig.\ \ref{fig:teaser} shows an example the ``universal camera ISP'' idea, where we render an iPhone raw image -- after mapping to the target sensor using our method -- through an ISP designed for rendering Samsung raw images.

Despite its importance, the majority of prior work focuses on colorimetric calibration\footnote{Colorimetric calibration maps camera colors to corresponding device-independent tristimulus values in some canonical space (typically the CIE XYZ space).} (e.g., \cite{hong2001study, 7508874, finlayson2017color, 7047834}). There is a lack of prior work related to raw-to-raw mapping. Nguyen et al., \cite{nguyen2014raw}, to the best of our knowledge, proposed the first attempt for raw-to-raw mapping and showed that classical color mapping approaches, originally proposed for colorimetric calibration, can be used for raw-to-raw mapping. Specifically, Nguyen et al., \cite{nguyen2014raw} proposed to compute a pairwise raw-to-raw calibration to map image $I_A$ to image $I_B$, achieving very promising results. This mapping can be expressed as follows:
\begin{equation}
\label{eq1}
\hat{I}_B =g\left(M\texttt{ }\phi\left(r\left(I_A\right)\right)\right),
\end{equation}
where $r\left(\cdot\right)$ and $g\left(\cdot\right)$ are reshaping functions that represent images as $3\!\times\!n$ and $h\!\times\!w\times\!3$, respectively. $n = hw$ is the total number of pixels in each image, $M$ is a color mapping matrix, and $\phi\left(\cdot\right)$ is a kernel function. To compute $M$, Nguyen et al. \cite{nguyen2014raw} used color calibration charts captured in each scene and computed a scene-specific mapping matrix. The work in \cite{nguyen2014raw} studied different ways to compute this mapping, including polynomial and identity kernel functions. Despite being accurate, the applicability of such pairwise calibration methods in real scenarios is limited as it requires capturing and annotating a calibration object for each new scene. Moreover, the color chart used for mapping has a limited number of color samples (i.e., the typical 24-color checker chart).

\paragraph{Contribution}
This paper discusses a practical raw-to-raw mapping with an easy setup to use. In particular, we propose a semi-supervised training to learn, from a very small set of paired images, a reasonable raw-to-raw mapping that does not require computing per-scene calibration (see Fig.\ \ref{fig:teaser}). In addition to this small paired set, we exploit another set of unpaired images (i.e., requires minimal capturing effort with no annotation) captured by each camera device to improve our mapping. As collecting such unpaired set is abundantly available, we believe that our method is the first to propose a \textit{practical setup} for this problem. Through extensive evaluation, we show that our method achieves better results compared to other domain adaptation alternatives (e.g., \cite{zhu2017unpaired, park2020contrastive, yang2020fda}). To enable training and evaluation of our new approach, we have collected a new dataset of raw images captured by two different smartphone cameras, i.e., iPhone X and Samsung Galaxy S9.

\section{Methodology} \label{sec:method}

\begin{figure}[t]
\includegraphics[width=\textwidth]{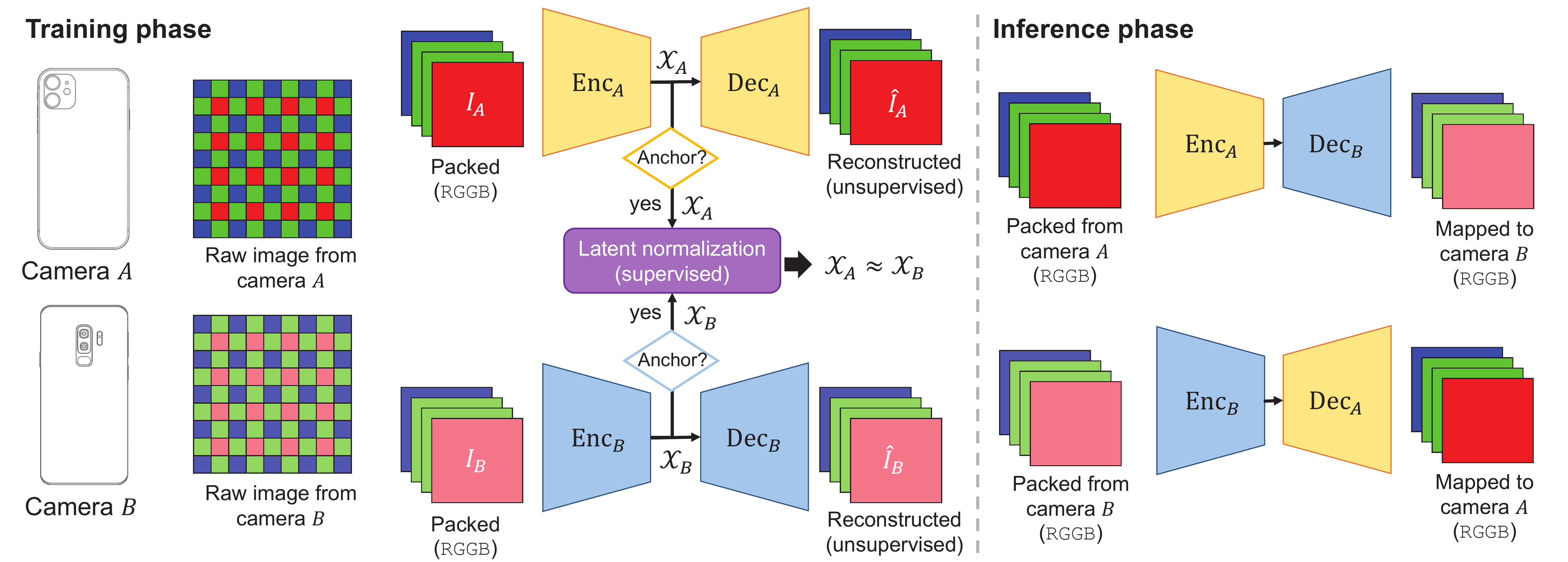}
\vspace{-1mm}
\caption{Our semi-supervised raw-to-raw mapping. In training, we use a set of unpaired images taken by $A$ and $B$ cameras, alongside a small set of paired images (so-called ``\textit{anchor set}''). We train two encoder-decoder networks to learn reconstructing the unpaired images from each camera. Additionally, we use our small anchor set to encourage the encoder net in both networks to produce similar latent features when input images share the same scene. At inference, we swap the decoder nets to map between raw spaces.\label{fig:main}}
\end{figure}

The overview of our method is shown in Fig.\ \ref{fig:main}. As illustrated, we propose to map raw images from camera $A$ to camera $B$ through a deep learning framework. Our framework includes two encoder-decoder networks, each of which is dedicated to one of our cameras. We use two different training sets: (i) a small set of paired images, the so-called ``anchor set'' and (ii) a larger set of unpaired images taken by cameras $A$ and $B$. 

At each iteration, we optimize both networks using data taken from each of these two sets. We penalize our network on the standard encoder-decoder reconstruction task when training data is from the unpaired set. In contrast, when training data is from the anchor set, we apply a latent normalization step to encourage the encoder of both networks to produce similar latent representations for each pair of images taken by our two different cameras, $A$ and $B$. The details of our loss functions are elaborated in Sec.\ \ref{subsec:method-loss}. At the inference phase, we swap our decoders to map images taken by camera $A$ to the raw space of camera $B$ and vice versa.

We use U-Net-like design \cite{unet} for each network that includes skip connections from different encoder blocks to the corresponding decoder blocks to improve the reconstruction accuracy. We used the same number of channels used in \cite{afifi2020deep} for our networks, as well as the inference post-processing used in \cite{afifi2020deep}.

\subsection{Anchor Set} \label{subsec:method-paired}
The first step necessary for our training is to generate the anchor set. This paired anchor set must include perfectly aligned images taken by both cameras $A$ and $B$. This is hard to achieve due to the drastic difference in optics and field of view between the two cameras. Even with careful capturing settings (e.g., \cite{ignatov2020replacing}), perfect alignment is not guaranteed. Due to this reason, and as our interest is in the color mapping (neither to map image noise characteristics nor quality), we adopted the standard pairwise calibration approach proposed in \cite{nguyen2014raw} to generate our paired set.

Specifically, we capture paired unaligned images by our two cameras with a color calibration chart in each scene. Then, we manually select corresponding colors from the color chart patches taken by each camera and use these colors to compute a polynomial mapping matrix, $M$, to map from camera $A$ to camera $B$. Afterward, we apply the computed matrix $M$ to all pixels produced by camera $A$ using Eq.\ \ref{eq1} to get aligned corresponding pixels in the raw space of camera $B$. Similarly, we map images by camera $B$ to the raw space of camera $A$. 

\begin{figure}[t]
\includegraphics[width=\textwidth]{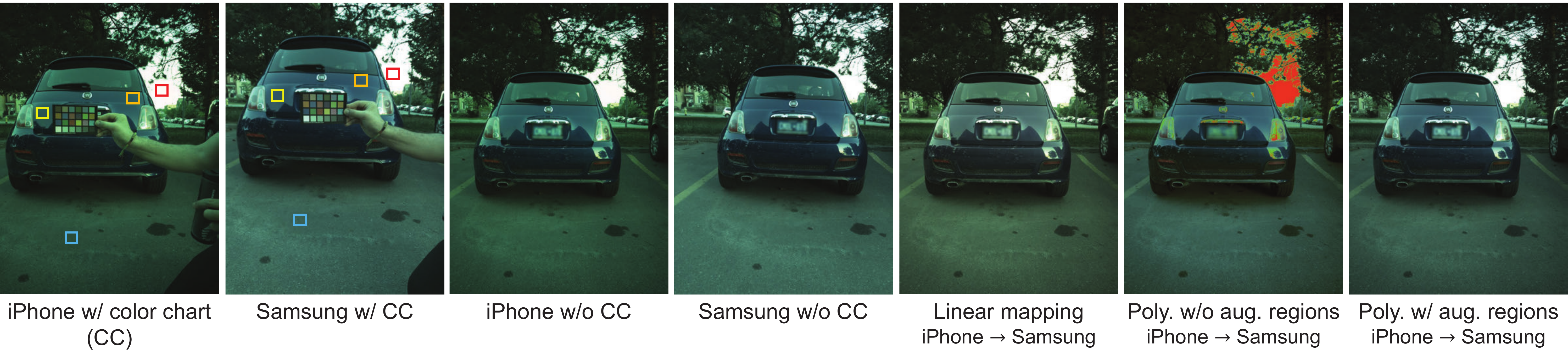}
\vspace{-1mm}
\caption{We propose to select other corresponding regions, in addition to the color chart's patches, to achieve better pairwise image calibration. As a comparison, we show calibration results using linear and polynomial mapping matrices w/ and w/o augmenting regions.\label{fig:dataset-mapping}}
\end{figure}

When computing $M$, we found that relying solely on the colors of chart patches does not always produce accurate mapping and may result in noticeable out-of-gamut mapping. This is due to the fact that, in some cases, the polynomial transformations tend to overfit to colors in the calibration chart and produces inaccurate mapping for other colors. To fix this, we propose to improve this mapping by the following. For each scene, we manually select corresponding homogeneous patches from other objects in each pair of images captured by camera $A$ and $B$ in order to improve the generalization of the polynomial fitting. To do that, we designed a tool to assist a human annotator to manually select patches by a simple drag and drop operation. In particular, we asked a human annotator to manually select a correspondence variable-size square patch from $A$ and $B$ such that the patches are strictly homogeneous. Our tool allows selecting a variable number of correspondence patches depending on the given image pair (i.e., more patches are better). We use these extra corresponding colors along with the colors of our calibration chart to compute $M$. This introduces a noticeable improvement in our mapping, as shown in Fig. \ref{fig:dataset-mapping}. Instead of applying the mapping to the image that contains the color chart, we capture another equivalent image (for each scene) without the color chart and apply the mapping to this chart-free image, as shown in Fig. \ref{fig:dataset-mapping}.

\subsection{Loss Function} \label{subsec:method-loss}
To optimize our framework, we used the following loss functions: (i) reconstruction loss, (ii) anchor loss, and (iii) mapping loss. The reconstruction loss $\reclossEQ$ is described as follows:

\begin{equation} \label{eq:rec-loss}
\reclossEQ = \frac{1}{N} \sum_{n=1}^{N} \left \| I_{n}  -  \hat{I}_{n}\right  \|_F^2,
\end{equation}

\noindent where $n$ is the mini-batch index, $I$ and $\hat{I}$ are the input and reconstructed image, respectively, and $\left \| \cdot  \right \|_F^2$ is squared Frobenius norm. Our mini-batch is designed to contain examples randomly from both unpaired and paired sets. During the training iteration, this loss is used for the samples from the unpaired set. For the samples from the anchor set, we encourage our encoders in both networks to normalize the latent features by using our anchor loss $\anchorlossEQ$:

\begin{equation} \label{eq:anch-loss}
\anchorlossEQ = \frac{1}{N} \sum_{e=1}^{E} \sum_{n=1}^{N} \left \| \mathcal{X}^{e}_{A_{n}}  -  \mathcal{X}^{e}_{B_{n}}\right  \|_F^2,
\end{equation}

\noindent where $e$ refers to encoder block indices; $\mathcal{X}^{e}_{A}$ and $\mathcal{X}^{e}_{B}$ are the latent output of the $e^\text{th}$ encoder block for camera $A$'s image and camera $B$'s image, respectively. As shown in Eq. \ref{eq:anch-loss}, instead of penalizing only on the output of the last block of our encoder, we aggregate the loss overall encoder blocks. This enables us to pass high-level latent representations from the encoder net to the decoder net through skip connections to help our reconstruction process. In addition to the anchor loss, we also incorporate our mapping loss:

\begin{equation} \label{eq:mapping-loss}
\mappinglossEQ = \frac{1}{2N} \sum_{n=1}^{N} \left \|  I_{A_{n}} - \hat{I}_{A_{n}} \right \|_F^2 + \left \|  I_{B_{n}} - \hat{I}_{B_{n}} \right \|_F^2,
\end{equation}

\noindent where $\hat{I}_{A_{n}}$, $I_{A_{n}}$, $\hat{I}_{B_{n}}$, and $I_{B_{n}}$ are the mapped and corresponding \textit{paired} ground-truth images by cameras $A$ and $B$, respectively. Our final loss function is then computed as:

\begin{equation} \label{eq:final-loss}
\mathcal{L} = \reclossEQ + \anchorlossEQ + \mappinglossEQ
\end{equation}

\section{Experiments and Discussion} \label{sec:results}

\subsection{Datasets} 
In this section, we provide a description of the datasets used in our proposed framework. Particularly, we use two datasets: (i) the NUS dataset \cite{cheng2014illuminant}, and (ii) our collected dataset. 

\paragraph{NUS Dataset} We used two sets captured by Canon EOS 600D and Nikon D5200 DSLR cameras. Note that although this dataset provides the same set of scenes, images in this dataset are not aligned and unpaired (i.e., the $i^{\text{th}}$ image in the Canon's set does not capture the same scene of the $i^{\text{th}}$ image in the Nikon's set). Thus, we manually selected seventeen corresponding images that capture the same scenes from each camera for testing. Additionally, we manually selected five corresponding images from each camera set as our anchor image set. We followed the same procedure explained in Sec.\ \ref{subsec:method-paired} to get aligned pairs for both test and anchor image sets. Thus, our final test set includes $34$ images for each camera, and our anchor set includes ten images for each camera---the number of images is doubled after mapping all images in each camera set to the other camera. We used the rest of the unpaired images in each camera set to construct our unpaired training sets. In particular, we used $172$ images in each camera set. The unpaired set along with the anchor set were used to train our method.

\begin{figure}[t]
\includegraphics[width=\textwidth]{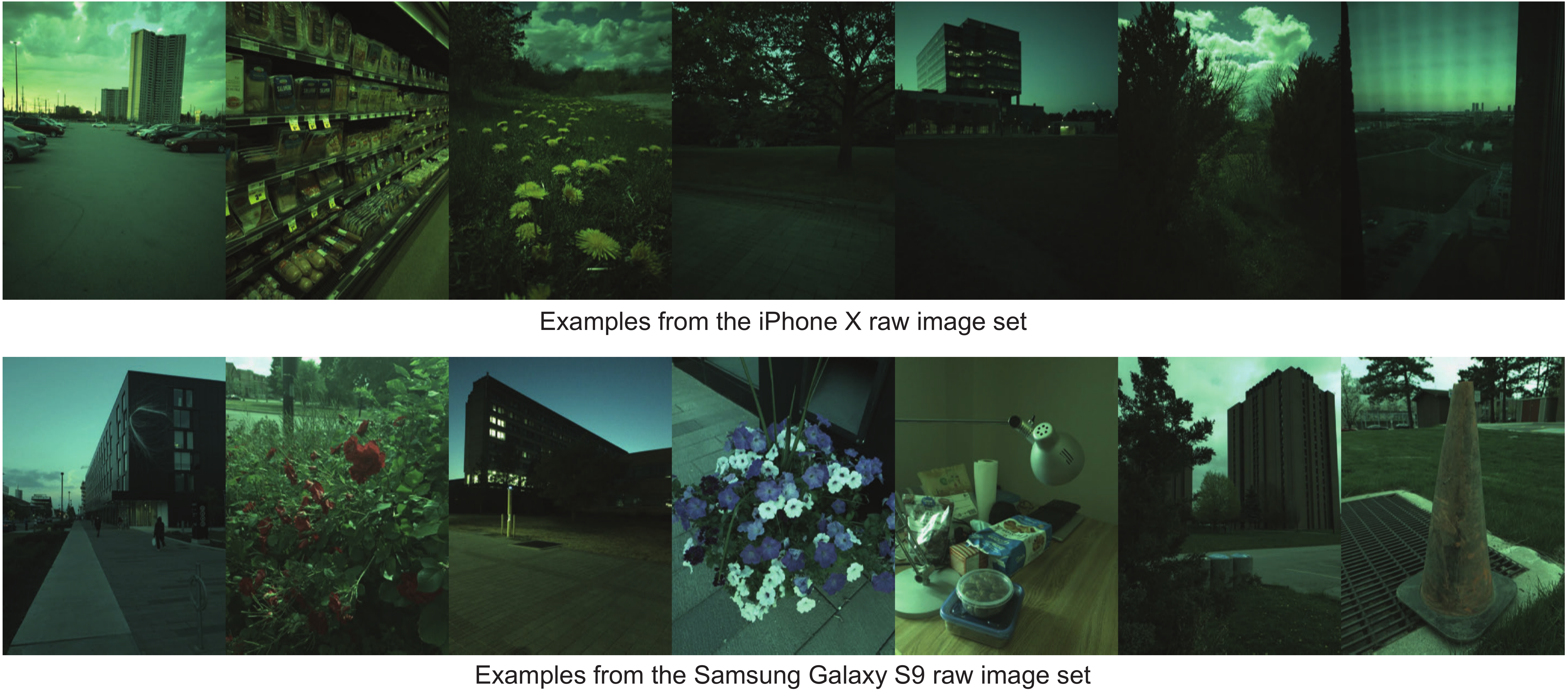}
\vspace{0.1mm}
\caption{Examples from each camera set. In the top row, we show examples from the iPhone X smartphone camera. In the bottom row, we show examples from the Samsung Galaxy S9 smartphone camera. To aid visualization, the Green channels are averaged (i.e., three-channel $\texttt{RGB}$ image), and a Gamma operation with 1/1.6 encoding gamma is applied.}
\label{fig:dataset-examples}
\end{figure}

\paragraph{Our Dataset} As one of our contributions, we propose a new dataset of raw images captured by two different smartphone cameras: Samsung Galaxy S9 and iPhone X. Fig.\ \ref{fig:dataset-examples} shows example raw images from each camera.  We opt in to use smartphone cameras for data collection due to the fact that smartphone cameras introduce large differences in spectral sensitivities compared to DSLR cameras~\cite{liba2019handheld}, which makes raw-to-raw mapping more challenging using smartphone cameras. As far as we know, there is no dataset of raw images captured by two or more smartphone cameras that meet our setup (i.e., contains unpaired and paired raw image sets).

Our dataset consists of an unpaired and paired set of images for each smartphone camera. The unpaired set includes $196$ images captured by each smartphone camera (total of $392$). The paired set includes $115$ pair of images used for testing (generated as described in Sec.\ \ref{subsec:method-paired}). In addition to this paired set, we have another small set of $22$ anchor paired images. Original DNG files of our raw images, in addition to the associated source code to extract raw images and metadata, are available online\footnote{\href{https://github.com/mahmoudnafifi/raw2raw}{https://github.com/mahmoudnafifi/raw2raw}}. To train our method, we used $x$ paired anchor images (we present an ablation study using $x = \{1, 7, 15, 22\}$ in Sec. \ref{subsec:results-comparisons}). We used the unpaired set for each camera for the unsupervised part of our training.  %Fig.\ \ref{fig:anchor-chromaticity} shows $rg$ chromaticity plots of colors in the unpaired and anchor training sets for each camera in our dataset. As shown, the colors produced by each camera sensor in our training set occupy a different region in the $rg$ chromaticity space and even for the paired anchor set, mapping between sensor colors is non-trivial. Fig.\ \ref{fig:anchor-chromaticity} also shows the CIE $xy$ chromaticity of the X-Rite color checker chart's colors captured under outdoor and indoor lighting conditions by each camera. It can be seen that the correlation between color correspondences in the color chart changes based on the lighting condition.

% \begin{figure}[t]
% \includegraphics[width=\textwidth]{images/anchor_chromaticity.png}
% \vspace{0.1mm}
% \caption{Top: $rg$ chromaticity of the entire training set (i.e., unpaired and anchor sets) for each camera in our dataset. Bottom: CIE $xy$ chromaticity of color charts captured under outdoor and indoor lighting conditions by each camera used in our dataset.}
% \label{fig:anchor-chromaticity}
% \end{figure}

\subsection{Implementation Details} 
All images are processed after black-level subtraction and image normalization. We examined demosaiced and mosaiced (Bayer) images in our experiments. For the experiments on the NUS dataset \cite{cheng2014illuminant}, we used sensor raw images that were minimally processed using the DCRAW converter as described in the dataset paper \cite{cheng2014illuminant}. Accordingly, the network's first and last layers were modified to accept/output three-channel images instead of the four channels shown in Fig.\ \ref{fig:main}. We used the metadata provided in the NUS dataset to get black and white levels. For the experiments in our dataset, we used $\texttt{RGGB}$ images after packing each mosaiced image into four channels. For each image, we extracted the black and white levels from the associated DNG file \cite{DNG}. We end-to-end optimized our network's weights using Adam algorithm \cite{kingma2014adam} with a learning rate of $10^{-4}$ and beta values $0.9$ and $0.999$. We set the mini-batch size $N$ to $16$, and we trained on $256\!\times\!256$ patches randomly selected from each training image for $140$ epochs.  

\subsection{Comparisons and Ablation Studies} \label{subsec:results-comparisons}

\begin{figure}[t]
\includegraphics[width=\textwidth]{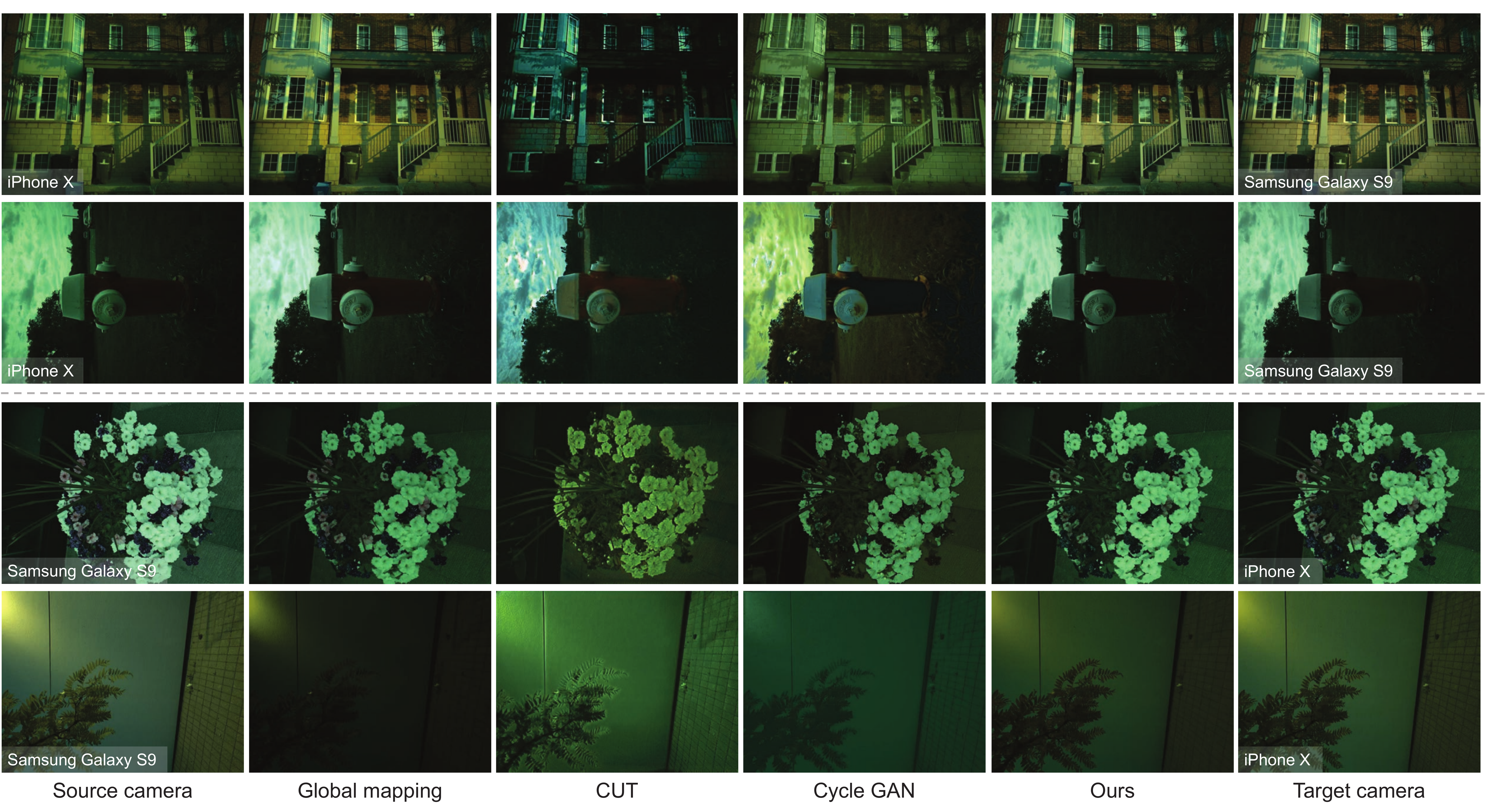}
\vspace{-1mm}
\caption{Qualitative comparisons on our dataset. We show our results alongside the results of: global mapping, CUT \cite{park2020contrastive}, and Cycle GAN \cite{zhu2017unpaired}. For each example, we show the ``ground-truth`` image in the target camera's raw space.}
\label{fig:qualitative}
\end{figure}

% NUS
\begin{table*}[t]
\caption{Results of mapping between  Canon EOS 600D and Nikon D5200 DSLR cameras from the NUS dataset \cite{cheng2014illuminant}. This table also provides the results of our ablation studies on the effect of our loss terms (\mappingloss, \anchorloss, and \recloss). Best results are highlighted in yellow and boldfaced. \label{table:results-NUS}}
\vspace{3mm}
\centering
\scalebox{0.59}{
\begin{tabular}{|l|c|c|c|c|c|c|c|c|}
\hline
 & \multicolumn{4}{c|}{\cellcolor[HTML]{DEFFC9}Canon $\rightarrow$ Nikon} & \multicolumn{4}{c|}{\cellcolor[HTML]{DEFFC9}Nikon $\rightarrow$ Canon} \\ \cline{2-9} 
\multirow{-2}{*}{Method} &  PSNR$\uparrow$ & SSIM$\uparrow$ & MAE$\downarrow$ & $\Delta$ E$\downarrow$ & PSNR$\uparrow$ & SSIM$\uparrow$ & MAE$\downarrow$ & $\Delta$ E$\downarrow$ \\  \hline

Global calibration ($3\!\times\!3$) & 26.99 $\pm$ 1.66 & 0.84 $\pm$ 0.04 & 0.042 $\pm$ 0.01 & 8.66 $\pm$ 1.89 &
27.84 $\pm$ 2.28 & 0.84 $\pm$ 0.04 & 0.04 $\pm$ 0.01 & 7.76 $\pm$ 1.87 \\ \hline
Global calibration (poly) & 21.08 $\pm$ 4.58 & 0.77 $\pm$  0.07 & 0.087 $\pm$ 0.05 & 12.57 $\pm$ 2.45 & 
22.43 $\pm$ 7.00 & 0.78 $\pm$ 0.09 &  0.10 $\pm$ 0.05 & 11.82 $\pm$ 4.21 \\ \hline
Cycle GAN \cite{zhu2017unpaired} &  27.33 & 0.83 & 0.030 & 15.31 &
26.85 & 0.83 &  0.031 & 11.49 \\ \hline
CUT \cite{park2020contrastive} & 26.06 & 0.71 & 0.040 & 16.24 & 
29.85 & 0.88 & 0.022 & 10.35 \\ \cdashline{1-9}

Ours (w/o \anchorloss and \recloss)   & 30.17 & 0.91 & 0.022 & 6.38 & 
29.69 & 0.91 & 0.024 & 6.23 \\ \hline
Ours (w/o \recloss)  & 29.94 & 0.91 & 0.023 & \cellcolor[HTML]{FFFFC7}\textbf{6.21} & 
30.19 & 0.90 & 0.024 & 6.10\\ \hline
Ours (w/o \anchorloss)  & 29.85 & 0.90 & 0.023 & 6.27
& 29.10 & 0.89 & 0.025 & 6.03\\ \hline
Ours (w/o \mappingloss)  & 31.15 & 0.91 & 0.022 & 7.01 & 
29.81 & 0.91 & 0.026 & 6.67 \\ \hline
Ours & \cellcolor[HTML]{FFFFC7}\textbf{32.36} & \cellcolor[HTML]{FFFFC7}\textbf{0.93} & \cellcolor[HTML]{FFFFC7}\textbf{0.020} & \cellcolor[HTML]{FFFFC7}\textbf{6.21}  &
\cellcolor[HTML]{FFFFC7}\textbf{30.81} & \cellcolor[HTML]{FFFFC7}\textbf{0.93} & \cellcolor[HTML]{FFFFC7}\textbf{0.023}  & \cellcolor[HTML]{FFFFC7}\textbf{5.95} \\ \hline
\end{tabular}}
\end{table*}

% Our dataset
\begin{table*}[t]
\caption{Results of mapping between Samsung Galaxy S9 and iPhone X cameras on our dataset. The symbol $x$ refers to the number of paired raw images used in the anchor set. Best results are highlighted in yellow and boldfaced. \label{table:results-ourdataset}}
\vspace{3mm}
\centering
\scalebox{0.58}{
\begin{tabular}{|l|c|c|c|c|c|c|c|c|}
\hline
 & \multicolumn{4}{c|}{\cellcolor[HTML]{DEFFC9}Samsung $\rightarrow$ iPhone} & \multicolumn{4}{c|}{\cellcolor[HTML]{DEFFC9}iPhone $\rightarrow$ Samsung} \\ \cline{2-9} 
\multirow{-2}{*}{Method} &  PSNR$\uparrow$ & SSIM$\uparrow$ & MAE$\downarrow$ & $\Delta$ E$\downarrow$ &  PSNR$\uparrow$ & SSIM$\uparrow$ & MAE$\downarrow$ & $\Delta$ E$\downarrow$ \\  \hline

Global calibration ($3\!\times\!3$)  & 24.52 $\pm$ 3.13 & 0.71 $\pm$ 0.16 & 0.049 $\pm$ 0.02 & 10.22 $\pm$ 3.60 & 
17.03 $\pm$ 7.13 & 0.51 $\pm$ 0.30 & 0.16 $\pm$ 0.12 & 18.76 $\pm$ 11.32 \\ \hline
Global calibration (poly) & 24.88 $\pm$ 2.82 & 0.72 $\pm$ 0.16 & 0.048 $\pm$ 0.02 & 10.08 $\pm$ 3.71 & 
16.88 $\pm$ 7.18 & 0.50 $\pm$ 0.30 & 0.16 $\pm$ 0.12 & 19.26 $\pm$ 11.23 \\ \hline
FDA \cite{yang2020fda}  & 20.95 $\pm$ 0.28 & 0.48 $\pm$ 0.03 & 0.06 $\pm$ 0.002 & 21.31 $\pm$ 0.86 & 
19.18 $\pm$ 0.50 & 0.47 $\pm$ 0.02 & 0.090 $\pm$ 0.004 & 22.55 $\pm$ 1.09 \\ \hline
Cycle GAN \cite{zhu2017unpaired} & 24.52 & 0.71 &  0.043 & 14.64 & 
24.35 & 0.75 & 0.043 & 12.84 \\ \hline
CUT \cite{park2020contrastive} & 22.24 & 0.71 & 0.051 &  15.16 & 
22.79 & 0.74 & 0.046 & 14.94 \\ \cdashline{1-9}

Ours ($x=1$)  & 28.42 & 0.87 & 0.031 & \cellcolor[HTML]{FFFFC7}\textbf{5.95} & 
26.52 & 0.86 & 0.039 & 6.95 \\ \hline
Ours ($x=7$)  & 28.64 & 0.88 & 0.028 & 6.21 & 26.70 & 0.87 & 0.039 & 6.93 \\ \hline
Ours ($x=15$)  & 29.24 & 0.88 & \cellcolor[HTML]{FFFFC7}\textbf{0.027} & 6.66 & 
\cellcolor[HTML]{FFFFC7}\textbf{28.59} & \cellcolor[HTML]{FFFFC7}\textbf{0.90} & \cellcolor[HTML]{FFFFC7} \textbf{0.032} & 6.61 \\ \hline
Ours ($x=22$)  & \cellcolor[HTML]{FFFFC7}\textbf{29.65} & \cellcolor[HTML]{FFFFC7}\textbf{0.89} & \cellcolor[HTML]{FFFFC7}\textbf{0.027} & 6.32 & 
28.58 & \cellcolor[HTML]{FFFFC7}\textbf{0.90} & 0.033 & \cellcolor[HTML]{FFFFC7}\textbf{6.53} \\ \hline

\end{tabular}}
\end{table*}

We compared our method with global calibration methods used for raw-to-raw mapping \cite{nguyen2014raw}. Specifically, we compared our method with two different global calibration approaches: (i) the linear $3\!\times\!3$ mapping and (ii) the polynomial mapping (here, we used the same polynomial function used in \cite{hong2001study, nguyen2014raw}). We first compute a single (either $3\!\times\!3$ or polynomial) calibration matrix from a single pair of color charts captured by cameras $A$ and $B$. Then, we applied the computed calibration matrix to the testing images. We repeated this experiment multiple times to avoid any bias to the selected pair of images using all pairs in the anchor set. We reported the mean and standard deviation of the results. We further examined the recent Fourier domain adaptation (FDA) method proposed in \cite{yang2020fda}. To test the FDA method \cite{yang2020fda}, we followed a similar setting, where we used one of the anchor images as a target image and apply the FDA between this image and all testing images. We repeated this process overall anchor images and reported the mean and standard deviation of the results.

\begin{figure}[t]
\includegraphics[width=\textwidth]{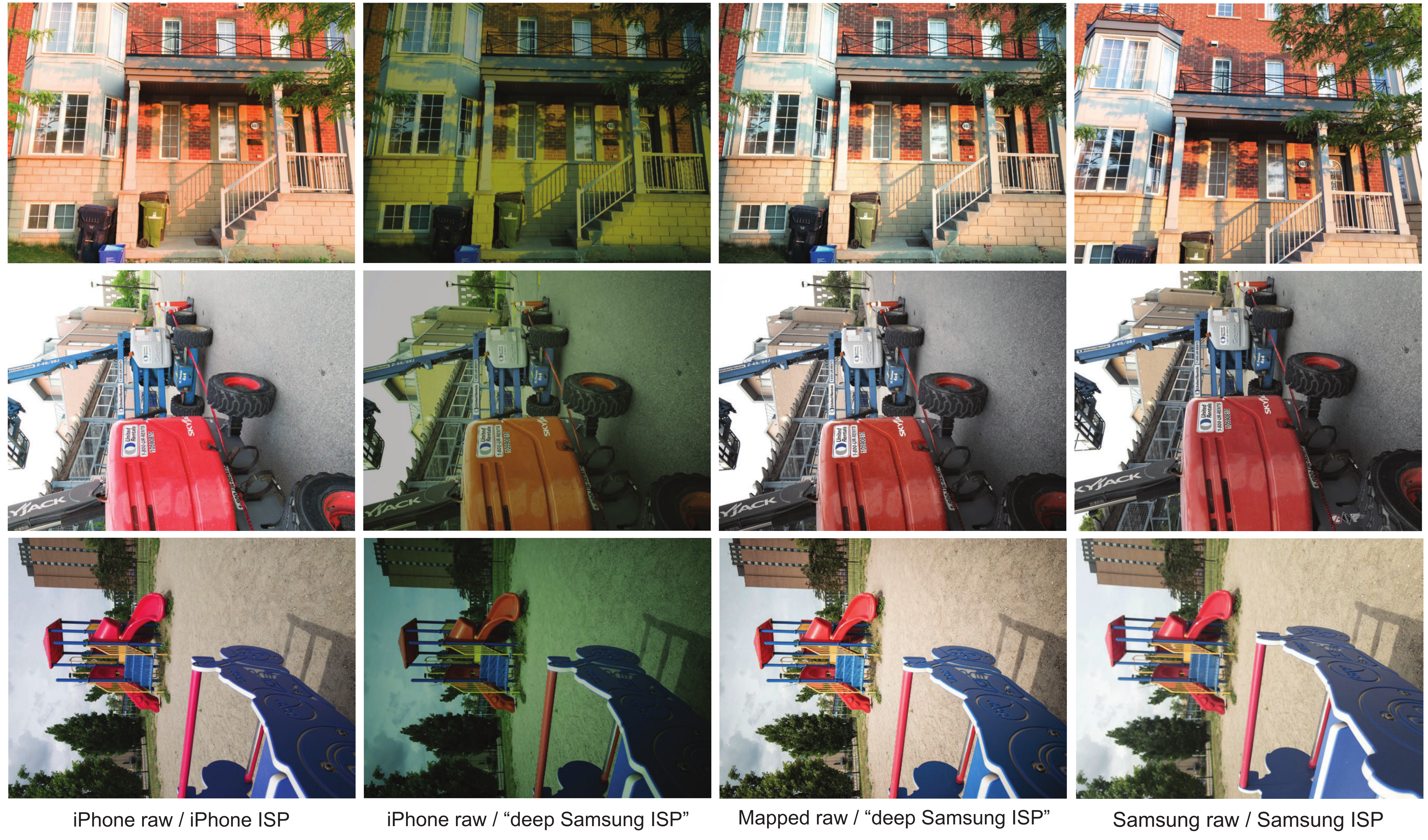}
\vspace{-1mm}
\caption{This figure shows example sRGB images captured and rendered by two different cameras (iPhone [$1^\text{st}$ column] and Samsung [$4^\text{th}$ column]). The $2^\text{nd}$ and $3^\text{rd}$ columns show results of a deep learning-based camera ISP model trained on Samsung raw/sRGB images. We show examples of feeding iPhone raw images to this trained deep ISP in the $2^\text{nd}$ column. We also show in the $3^\text{rd}$ column the results of rendering the raw images (mapped from iPhone to Samsung using our raw-to-raw mapping) through the trained Samsung ISP.}
\label{fig:rendering}
\end{figure}

\begin{figure}[t]
\includegraphics[width=\textwidth]{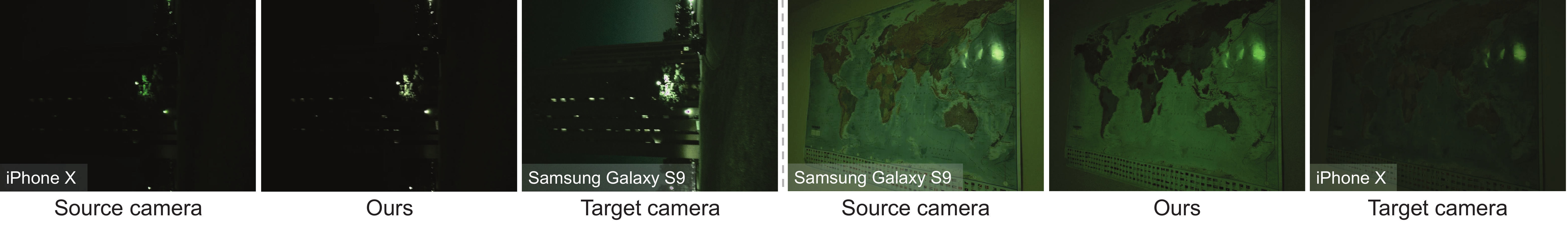}
\vspace{-1mm}
\caption{Failure cases. Our method fails in some cases, especially with dark scenes or scenes with challenging light conditions.}
\label{fig:faliure}
\vspace{-2mm}
\end{figure}

In addition, we compared our method with two unsupervised image-to-image translation methods. In particular, we reported the results of the Cycle GAN method \cite{zhu2017unpaired} and the contrastive unpaired translation (CUT) method \cite{park2020contrastive} after training the network of each method using the same training settings used to train our network, with the exception that both Cycle GAN and CUT networks were found to require more epochs to converge. Thus, we increased the training duration to 450 epochs ($\sim$3 times our training epochs). When reporting results on packed Bayer images, the network architectures of Cycle GAN \cite{zhu2017unpaired} and the CUT \cite{park2020contrastive} methods were modified to accept and output four-channel images.

We also conducted a set of ablation studies to evaluate the improvement gained by our introduced loss terms (i.e., \mappingloss, \anchorloss, and \recloss). Note that when using only the mapping loss term, \mappingloss, this is equivalent to train two separate networks on the paired anchor set.

Results of our method, including our ablation studies, and competitive methods on the NUS dataset \cite{nguyen2014raw} are reported in Table \ref{table:results-NUS}. We show the PSNR score, SSIM score \cite{wang2004image}, and mean absolute error (MAE) achieved by each method. Additionally, we measure the visual perception errors of each method using $\Delta$ E 2000 \cite{sharma2005ciede2000}. As computing $\Delta$ E is performed in the CIE Lab space, we first transform  each of the mapped raw images and corresponding ground-truth images to the CIE XYZ space, then we mapped the XYZ images to the CIE Lab space. Mapping images to the CIE XYZ space was performed using the calibration matrices provided by each camera manufacturer. The illuminant vectors -- which are required for the CIE XYZ mapping \cite{karaimer2018improving, afifi2020cross} -- of the mapped images were obtained using the state-of-the-art sensor-independent illuminant estimation method \cite{afifi2020cross}, while we used the achromatic patches in the color calibration chart to determine the illuminant vectors of ground-truth images.

Table \ref{table:results-ourdataset} shows the results on our proposed dataset. In Table \ref{table:results-ourdataset}, we also report the results of using different number of images in the anchor set ($x = \{1, 7, 15, 22\}$). As shown, even when using a single pair of anchor images, our method achieves superior results to other competitors. We acknowledge that studying the number of anchor images does not give enough insights about the characteristics of anchor data to improve the results (in comparison to, for example, studying the impact of lighting conditions of the anchor set); however, one could study this research question in future work. Note that when we used $x<22$ in the ablation study, we randomly selected the anchor images from our anchor set, which includes 22 images for each camera.

We provide qualitative comparisons in Fig.\ \ref{fig:qualitative}. Additionally, we examined the idea of ``universal camera ISP'' mentioned in Sec.\ \ref{sec:intro}. Specifically, we trained a deep neural network on our Samsung S9 unpaired set to learn rendering raw images to the sRGB color space. We adopted the same architecture and training setting of AWNet \cite{dai2020awnet}. Then, we feed raw images from iPhone X with and without our raw-to-raw mapping to this trained deep ISP. As shown in Fig.\ \ref{fig:rendering}, our mapped raw gives reasonable sRGB that perceptually look similar to the actual output from Samsung S9.

\subsection{Limitations and Future Work} \label{sec:subsec:results-limitations}
Given the practical setup used in our experiments, we achieve state-of-the-art results compared to other alternatives. However, our results still need more improvement to be able to achieve more robust raw-to-raw mapping. Specifically, our method fails in some cases, as shown in Fig.\ \ref{fig:faliure}, where it could not produce good results in dark scenes or scenes with challenging lighting conditions.

Capturing conditions and settings can provide useful cues for better raw-to-raw mapping. To further guide the training process, one could incorporate raw-image metadata (e.g., exposure time, ISO, noise level, and focus distance) to improve the accuracy and robustness. Another potential improvement could be achieved by considering an adaptive and sensor-specific amplification for low-light scenes.

\section{Conclusion} \label{sec:conclusion}
We have presented a semi-supervised raw-to-raw mapping method. Our work presents a practical way to achieve this mapping with a limited set of paired images required to train the model. Under this practical scenario, we demonstrated state-of-the-art results on two different datasets of DSLR and a new proposed smartphone camera dataset. Our method is the first step towards having practical and accurate raw-to-raw mapping to assist camera ISP manufacturing.

\bibliography{ref}

\begin{thebibliography}{27}
\providecommand{\natexlab}[1]{#1}
\providecommand{\url}[1]{\texttt{#1}}
\expandafter\ifx\csname urlstyle\endcsname\relax
  \providecommand{\doi}[1]{doi: #1}\else
  \providecommand{\doi}{doi: \begingroup \urlstyle{rm}\Url}\fi

\bibitem[DNG(2012)]{DNG}
Digital negative ({D}{N}{G}) specification.
\newblock Technical report, Adobe Systems Incorporated, 2012.
\newblock Version 1.4.0.0.

\bibitem[Afifi and Brown(2020)]{afifi2020deep}
Mahmoud Afifi and Michael~S Brown.
\newblock Deep white-balance editing.
\newblock In \emph{CVPR}, 2020.

\bibitem[Afifi et~al.(2021{\natexlab{a}})Afifi, Abdelhamed, Abuolaim,
  Punnappurath, and Brown]{afifi2021ciexyz}
Mahmoud Afifi, Abdelrahman Abdelhamed, Abdullah Abuolaim, Abhijith
  Punnappurath, and Michael~S Brown.
\newblock {CIE} {XYZ} {N}et: Unprocessing images for low-level computer vision
  tasks.
\newblock \emph{IEEE Transactions on Pattern Analysis and Machine
  Intelligence}, pages 1--12, 2021{\natexlab{a}}.

\bibitem[Afifi et~al.(2021{\natexlab{b}})Afifi, Barron, LeGendre, Tsai, and
  Bleibel]{afifi2020cross}
Mahmoud Afifi, Jonathan~T Barron, Chloe LeGendre, Yun-Ta Tsai, and Francois
  Bleibel.
\newblock Cross-camera convolutional color constancy.
\newblock In \emph{ICCV}, 2021{\natexlab{b}}.

\bibitem[Andersen and Connah(2016)]{7508874}
Casper~Find Andersen and David Connah.
\newblock Weighted constrained hue-plane preserving camera characterization.
\newblock \emph{IEEE Transactions on Image Processing}, 25\penalty0
  (9):\penalty0 4329--4339, 2016.

\bibitem[Basri and Jacobs(2003)]{basri2003lambertian}
Ronen Basri and David~W Jacobs.
\newblock Lambertian reflectance and linear subspaces.
\newblock \emph{IEEE Transactions on Pattern Analysis and Machine
  Intelligence}, 25\penalty0 (2):\penalty0 218--233, 2003.

\bibitem[Cheng et~al.(2014)Cheng, Prasad, and Brown]{cheng2014illuminant}
Dongliang Cheng, Dilip~K Prasad, and Michael~S Brown.
\newblock Illuminant estimation for color constancy: why spatial-domain methods
  work and the role of the color distribution.
\newblock \emph{Journal of the Optical Society of America A}, 31\penalty0
  (5):\penalty0 1049--1058, 2014.

\bibitem[Dai et~al.(2020)Dai, Liu, Li, and Chen]{dai2020awnet}
Linhui Dai, Xiaohong Liu, Chengqi Li, and Jun Chen.
\newblock {AWNet}: Attentive wavelet network for image isp.
\newblock \emph{arXiv preprint arXiv:2008.09228}, 2020.

\bibitem[Finlayson et~al.(2017)Finlayson, Gong, and Fisher]{finlayson2017color}
Graham Finlayson, Han Gong, and Robert~B Fisher.
\newblock Color homography: theory and applications.
\newblock \emph{IEEE Transactions on Pattern Analysis and Machine
  Intelligence}, 41\penalty0 (1):\penalty0 20--33, 2017.

\bibitem[Finlayson et~al.(2015)Finlayson, Mackiewicz, and Hurlbert]{7047834}
Graham~D. Finlayson, Michal Mackiewicz, and Anya Hurlbert.
\newblock Color correction using root-polynomial regression.
\newblock \emph{IEEE Transactions on Image Processing}, 24\penalty0
  (5):\penalty0 1460--1470, 2015.

\bibitem[Fourure et~al.(2016)Fourure, Emonet, Fromont, Muselet, Tr{\'e}meau,
  and Wolf]{fourure2016mixed}
Damien Fourure, R{\'e}mi Emonet, Elisa Fromont, Damien Muselet, Alain
  Tr{\'e}meau, and Christian Wolf.
\newblock Mixed pooling neural networks for color constancy.
\newblock In \emph{ICIP}, 2016.

\bibitem[Hong et~al.(2001)Hong, Luo, and Rhodes]{hong2001study}
Guowei Hong, M~Ronnier Luo, and Peter~A Rhodes.
\newblock A study of digital camera colorimetric characterization based on
  polynomial modeling.
\newblock \emph{Color Research \& Application}, 26\penalty0 (1):\penalty0
  76--84, 2001.

\bibitem[Ignatov et~al.(2020)Ignatov, Van~Gool, and
  Timofte]{ignatov2020replacing}
Andrey Ignatov, Luc Van~Gool, and Radu Timofte.
\newblock Replacing mobile camera {ISP} with a single deep learning model.
\newblock In \emph{CVPR Workshops}, 2020.

\bibitem[Jiang et~al.(2013)Jiang, Liu, Gu, and S{\"u}sstrunk]{jiang2013space}
Jun Jiang, Dengyu Liu, Jinwei Gu, and Sabine S{\"u}sstrunk.
\newblock What is the space of spectral sensitivity functions for digital color
  cameras?
\newblock In \emph{WACV}, 2013.

\bibitem[Karaimer and Brown(2018)]{karaimer2018improving}
Hakki~Can Karaimer and Michael~S Brown.
\newblock Improving color reproduction accuracy on cameras.
\newblock In \emph{CVPR}, 2018.

\bibitem[Kingma and Ba(2014)]{kingma2014adam}
Diederik~P Kingma and Jimmy Ba.
\newblock Adam: A method for stochastic optimization.
\newblock \emph{arXiv preprint arXiv:1412.6980}, 2014.

\bibitem[Liang et~al.(2021)Liang, Cai, Cao, and Zhang]{liang2021cameranet}
Zhetong Liang, Jianrui Cai, Zisheng Cao, and Lei Zhang.
\newblock Camera{N}et: A two-stage framework for effective camera {ISP}
  learning.
\newblock \emph{IEEE Transactions on Image Processing}, 30:\penalty0
  2248--2262, 2021.

\bibitem[Liba et~al.(2019)Liba, Murthy, Tsai, Brooks, Xue, Karnad, He, Barron,
  Sharlet, Geiss, et~al.]{liba2019handheld}
Orly Liba, Kiran Murthy, Yun-Ta Tsai, Tim Brooks, Tianfan Xue, Nikhil Karnad,
  Qiurui He, Jonathan~T Barron, Dillon Sharlet, Ryan Geiss, et~al.
\newblock Handheld mobile photography in very low light.
\newblock \emph{ACM Transactions on Graphics (TOG)}, 38\penalty0 (6):\penalty0
  1--16, 2019.

\bibitem[Lou et~al.(2015)Lou, Gevers, Hu, Lucassen, et~al.]{lou2015color}
Zhongyu Lou, Theo Gevers, Ninghang Hu, Marcel~P Lucassen, et~al.
\newblock Color constancy by deep learning.
\newblock In \emph{BMVC}, 2015.

\bibitem[Nguyen et~al.(2014)Nguyen, Prasad, and Brown]{nguyen2014raw}
Rang Nguyen, Dilip~K Prasad, and Michael~S Brown.
\newblock Raw-to-raw: Mapping between image sensor color responses.
\newblock In \emph{CVPR}, 2014.

\bibitem[Park et~al.(2020)Park, Efros, Zhang, and Zhu]{park2020contrastive}
Taesung Park, Alexei~A Efros, Richard Zhang, and Jun-Yan Zhu.
\newblock Contrastive learning for unpaired image-to-image translation.
\newblock In \emph{ECCV}, 2020.

\bibitem[Ronneberger et~al.(2015)Ronneberger, Fischer, and Brox]{unet}
Olaf Ronneberger, Philipp Fischer, and Thomas Brox.
\newblock {U}-{N}et: Convolutional networks for biomedical image segmentation.
\newblock In \emph{MICCAI}, 2015.

\bibitem[Schwartz et~al.(2018)Schwartz, Giryes, and
  Bronstein]{schwartz2018deepisp}
Eli Schwartz, Raja Giryes, and Alex~M Bronstein.
\newblock Deep{ISP}: Toward learning an end-to-end image processing pipeline.
\newblock \emph{IEEE Transactions on Image Processing}, 28\penalty0
  (2):\penalty0 912--923, 2018.

\bibitem[Sharma et~al.(2005)Sharma, Wu, and Dalal]{sharma2005ciede2000}
Gaurav Sharma, Wencheng Wu, and Edul~N Dalal.
\newblock The {CIEDE2000} color-difference formula: Implementation notes,
  supplementary test data, and mathematical observations.
\newblock \emph{Color Research \& Application}, 30\penalty0 (1):\penalty0
  21--30, 2005.

\bibitem[Wang et~al.(2004)Wang, Bovik, Sheikh, and Simoncelli]{wang2004image}
Zhou Wang, Alan~C Bovik, Hamid~R Sheikh, and Eero~P Simoncelli.
\newblock Image quality assessment: from error visibility to structural
  similarity.
\newblock \emph{IEEE Transactions on Image Processing}, 13\penalty0
  (4):\penalty0 600--612, 2004.

\bibitem[Yang and Soatto(2020)]{yang2020fda}
Yanchao Yang and Stefano Soatto.
\newblock {FDA}: Fourier domain adaptation for semantic segmentation.
\newblock In \emph{CVPR}, 2020.

\bibitem[Zhu et~al.(2017)Zhu, Park, Isola, and Efros]{zhu2017unpaired}
Jun-Yan Zhu, Taesung Park, Phillip Isola, and Alexei~A Efros.
\newblock Unpaired image-to-image translation using cycle-consistent
  adversarial networks.
\newblock In \emph{ICCV}, 2017.

\end{thebibliography}
\end{document}